\pdfoutput=1

\documentclass[11pt]{article}

\usepackage[final]{acl}

\usepackage{times}
\usepackage{latexsym}
\usepackage{booktabs}
\usepackage{comment}
\usepackage{amsmath}
\usepackage{algorithm}
\usepackage{algpseudocode}
\usepackage{adjustbox}
\usepackage{url}

\usepackage[T1]{fontenc}

\usepackage[utf8]{inputenc}

\usepackage{microtype}

\usepackage{inconsolata}

\usepackage{graphicx}

\title{Multimodal Fine-grained Context Interaction Graph Modeling for Conversational Speech Synthesis}

\author{
  Zhenqi Jia\textsuperscript{1},
  Rui Liu\textsuperscript{1}\thanks{Corresponding author.},
  Berrak Sisman\textsuperscript{2},
  Haizhou Li\textsuperscript{3} \\
  \textsuperscript{1}Inner Mongolia University, Hohhot, China \\
  \textsuperscript{2}Center for Language and Speech Processing (CLSP), Johns Hopkins University, USA \\
  \textsuperscript{3}School of Artificial Intelligence, The Chinese University of Hong Kong, Shenzhen, China \\
  \texttt{jiazhenqi7@163.com, imucslr@imu.edu.cn, sisman@jhu.edu, haizhouli@cuhk.edu.cn}
}

\begin{document}
\maketitle

\begin{abstract}
Conversational Speech Synthesis (CSS) aims to generate speech with natural prosody by understanding the multimodal dialogue history (MDH). The latest work predicts the accurate prosody expression of the target utterance by modeling the utterance-level interaction characteristics of MDH and the target utterance. However, MDH contains fine-grained semantic and prosody knowledge at the word level. Existing methods overlook the fine-grained semantic and prosodic interaction modeling. To address this gap, we propose MFCIG-CSS, a novel \textbf{M}ultimodal \textbf{F}ine-grained \textbf{C}ontext \textbf{I}nteraction \textbf{G}raph-based CSS system. Our approach constructs two specialized multimodal fine-grained dialogue interaction graphs: a \textit{semantic interaction graph} and a \textit{prosody interaction graph}. These two interaction graphs effectively encode interactions between word-level semantics, prosody, and their influence on subsequent utterances in MDH. The encoded interaction features are then leveraged to enhance synthesized speech with natural conversational prosody. Experiments on the DailyTalk dataset demonstrate that MFCIG-CSS outperforms all baseline models in terms of prosodic expressiveness. Code and speech samples are available at https://github.com/AI-S2-Lab/MFCIG-CSS.
\end{abstract}

\section{Introduction}

Conversational speech synthesis (CSS) systems are required to generate speech with conversational interaction prosody, unlike traditional text-to-speech (TTS) systems \cite{guo2021conversational, liu2024generative, liu2024emphasis, liu2025retrieval}. With advances in user-agent interaction, CSS plays a key role in intelligent systems such as smartphone assistants \cite{vu2024gptvoicetasker}, smart homes \cite{jenal2022smart}, and virtual reality \cite{el2019virtual}.

Previous CSS methods improve prosody by modeling multimodal dialogue history (MDH) with coarse- and fine-grained context encoders \cite{lee2023dailytalk, hu2024fctalker, xue2023m, deng2024concss, li2022inferring}. However, they model coarse- and fine-grained features separately and overlook the interactive influence of word-level semantics and prosody on subsequent utterances. Additionally, some approaches \cite{li2022enhancing, liu2024emotion, jia2024intra} enhance prosody via speaking styles and emotional knowledge, but only consider utterance-level interactions, ignoring word-level effects.

The word-level semantics and prosody of key words in MDH play a crucial role in conversational interactions, directly influencing the semantics and prosody of subsequent utterances \cite{xue2023m, lin-etal-2024-advancing, castro-etal-2019-towards, li2024four, li2023imf, 9747565}. For example, in a conversation, when the user says ``I lost my wallet" and ``I lost my pen," they receive different responses in terms of both semantics and emotional prosody: a concerned inquiry ``Was there anything valuable in the wallet?" versus a relaxed inquiry ``Which pen did you lose?" The reason for the different responses is that the semantics expressed by the key words ``wallet" and ``pen" are different, and the user's emotional expression when saying these two words also differ.
Neglecting this word-level interaction modeling would limit the agent's ability to accurately capture semantic and prosodic variations in MDH, further affecting the modeling of the prosodic expressiveness of target utterance.  Therefore, how to model the interactions between word-level semantics, prosody, and the semantics, prosody of subsequent utterances in MDH to help the agent better understand the MDH and enhance the prosody expressiveness of the synthesized speech is the focus of this work.

\begin{figure*}[ht]
    \centering    \includegraphics[width=1\textwidth]{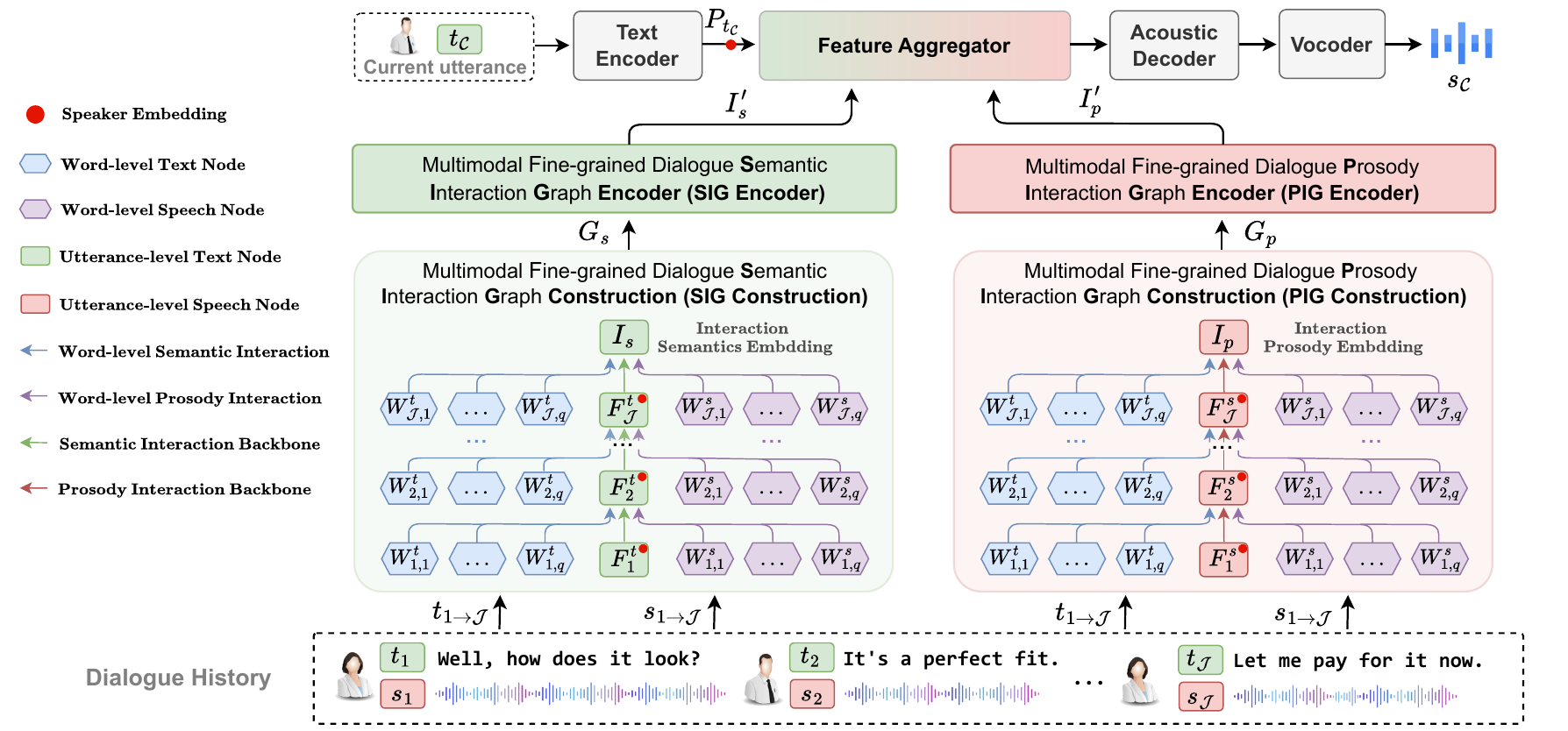}
    
    \caption{The overview of MFCIG-CSS consists of Multimodal Fine-grained Dialogue Semantic Interaction Graph, Multimodal Fine-grained Dialogue Prosody Interaction Graph, and Speech Synthesizer.}
    
    \label{fig2}
    
\end{figure*}

To address this issue, we propose a \textbf{M}ultimodal \textbf{F}ine-grained \textbf{C}ontext \textbf{I}nteraction \textbf{G}raph-based \textbf{CSS} system, termed \textbf{MFCIG-CSS}. Specifically, we design two multimodal fine-grained interaction graphs: a semantic interaction graph that models interactions between word-level semantics, prosody, and subsequent utterance semantics in MDH, and a prosody interaction graph that encodes interactions between word-level semantics, prosody, and subsequent utterance prosody. These interaction graphs enhance the agent’s understanding of MDH. Finally, we feed the encoded interaction features from both interaction graphs into the speech synthesizer to help the agent in synthesizing speech that aligns with conversational interaction prosody. In summary, the main contributions of this paper are as follows: 1) We propose MFCIG-CSS, a novel framework that models MDH interactions from the perspective of word-level semantics and prosody. 2) We design two interaction graphs—a semantic interaction graph and a prosody interaction graph—that explicitly capture and encode the fine-grained semantic and prosodic interactions in MDH, enhancing the system's contextual understanding. 3) Subjective and objective experiments on the DailyTalk dataset show that MFCIG-CSS outperforms all baseline models in terms of prosody expressiveness.

\section{Methodology}
\subsection{Task Definition}
A dialogue is defined as a sequence of utterances $\{[t_1, s_1], [t_2, s_2], ..., [t_{\mathcal{J}}, s_{\mathcal{J}}], [t_\mathcal{C}, s_\mathcal{C}]\}$, where $\{t_1, t_2, ..., t_{\mathcal{J}}\}$ represents the text of the dialogue history and $t_\mathcal{C}$ represents the text of the current utterance, $\{s_1, s_2, ..., s_{\mathcal{J}}\}$ represents the speech of the dialogue history and $s_\mathcal{C}$ represents the speech to be synthesized. For any $t_i$ and $s_i$,  \{${W^t_{i, 1}, \dots, W^t_{i, q}}$\} and \{${W^s_{i, 1}, \dots, W^s_{i, q}}$\} denote the word-level text and word-level speech of the $i$-th utterance, where $q$ denotes the number of words. 

\subsection{Model Overview}
The proposed MFCIG-CSS consists of three main components: 1) Multimodal Fine-grained Dialogue \textbf{S}emantic \textbf{I}nteraction \textbf{G}raph (\textbf{SIG}), 2) Multimodal Fine-grained Dialogue \textbf{P}rosody \textbf{I}nteraction \textbf{G}raph (\textbf{PIG}), and 3) Speech Synthesizer.
These modules will be described in detail in the following sections.

\subsection{SIG}
As shown on the left side of Figure \ref{fig2}, the SIG module consists of  \textbf{SIG Construction} and \textbf{SIG Encoder}. SIG captures the influence of word-level semantics and prosody on semantic interactions among subsequent utterances in dialogue history.

\textbf{SIG Construction.} To explicitly model the impact of word-level semantics and prosody on the subsequent utterance-level semantics interaction, we design an SIG $G_s = (\mathcal{N}, \mathcal{E})$, where $\mathcal{N}$ denotes the nodes and $\mathcal{E}$ denotes the relational edges between nodes. $G_s$ consists of three interaction branches, realized by three types of nodes (word-level text, word-level speech, and utterance-level text) and three types of relational edges. The three interaction branches are: \textbf{1) Word-level semantic interaction branch:} modeling the interaction between the word-level semantics and the subsequent utterance-level semantics in the dialogue history; \textbf{2) Word-level prosody interaction branch:} modeling the interaction between the word-level prosody and the subsequent utterance-level semantics in the dialogue history; \textbf{3) Semantic interaction backbone branch:} modeling the interaction between the utterance-level semantics and the subsequent utterance-level semantics in the dialogue history. Note that we add a special interaction semantic node ($I_s$) at the end of the semantic interaction backbone branch to integrate the interactive semantic features of the entire MDH. When initializing $G_s$, for the input $t_{1 \rightarrow \mathcal{J}}$, we use TOD-BERT \cite{wu-etal-2020-tod} to extract word-level text node features \{${W^t_{1, 1 \rightarrow q}, \dots, W^t_{\mathcal{J}, 1 \rightarrow q}}$\}, and use Sentence-BERT \cite{reimers-gurevych-2019-sentence} to extract utterance-level text node features \{${F^t_1, \dots, F^t_\mathcal{J}}$\}, while adding speaker embeddings to represent the identity of the speaker. For the input $s_{1 \rightarrow \mathcal{J}}$, we first use MFA to obtain each word's pronunciation segment, then use Wav2Vec2.0 \cite{baevski2020wav2vec} to extract frame-level
prosodic features and apply Average Pooling to obtain word-level speech node features \{${W^s_{1, 1 \rightarrow q}, \dots, W^s_{\mathcal{J}, 1 \rightarrow q}}$\}. We initialize $I_s$ with a zero vector.

\textbf{SIG Encoder.} We input the initialized $G_s$ into the SIG Encoder for encoding, learning the interaction of word-level semantics, prosody, and subsequent utterance-level semantics through three interaction branches. As shown in Equation (1), starting from the first sentence in the dialogue history, the utterance-level semantic feature ($F^t_i$), the word-level semantic features ($W^t_{i, 1 \rightarrow q}$), and the word-level prosody features ($W^s_{i, 1 \rightarrow q}$) of the i-th sentence are sequentially aggregated into the utterance-level semantic feature ($F^t_{i+1}$) of the (i+1)-th sentence. After all the nodes \{$F^t_{1}$, $F^t_{2}$, \dots, $F^t_{\mathcal{J}}$, $I_s$\} in the semantic interaction backbone branch fully interact with other word-level semantic and prosody nodes, we use Average Pooling to aggregate these interaction features into $I_s$, obtaining the final semantic interaction feature: $I'_s$.

\begin{equation}
\begin{aligned}
F^t_{i+1}  &= \text{SAGE}(F^t_{i}, W^t_{i, 1 \to q}, W^s_{i, 1 \to q}), \quad i \in [1, \mathcal{J}) \\
I_s &= \text{SAGE}(F^t_{\mathcal{J}}, W^t_{\mathcal{J}, 1 \to q}, W^s_{\mathcal{J}, 1 \to q}) \\
I'_s &= \text{Average Pooling}(F^t_{1 \to \mathcal{J}}, I_s)
\end{aligned}
\end{equation} 
where SAGE \cite{hamilton2017inductive}  denotes the graph convolution encoder.

\subsection{PIG}
As shown on the right side of Figure \ref{fig2}, the PIG module, similar to the SIG module, consists of \textbf{PIG Construction} and \textbf{PIG Encoder}. 
It is designed to model the influence of word-level semantics and prosody on the prosodic interactions of subsequent utterances in the dialogue history.

\textbf{PIG Construction.} We design an PIG $G_p$ in a similar structure to $G_s$. Note that the difference in construction compared to $G_s$ is that the third interaction branch of $G_p$ is the \textbf{prosody interaction backbone branch}, and at the end of this branch, a special interaction prosody node ($I_p$) is added to integrate the overall prosodic interaction features of MDH. During the initialization of $G_p$, we use Wav2Vec2.0-IEMOCAP\footnote{\label{wav2vec2iemocap}https://huggingface.co/speechbrain/emotion-recognition-wav2vec2-IEMOCAP} to extract utterance-level speech nodes and $I_p$ is initialized with a zero vector.

\textbf{PIG Encoder.} For the initialized $G_p$, we use the same architecture as the SIG Encoder to encode the interaction features between word-level semantics, prosody, and prosody of subsequent utterances in MDH, as shown in Equation (2). Finally, we apply Average Pooling to the interaction features \{$F^s_{1}$, $F^s_{2}$, \dots, $F^s_{\mathcal{J}}$, $I_p$\} to obtain the final prosodic interaction features: $I'_p$.

\begin{equation}
\begin{aligned}
F^s_{i+1}  &= \text{SAGE}(F^s_{i}, W^t_{i, 1 \to q}, W^s_{i, 1 \to q}), \quad i \in [1, \mathcal{J}) \\
I_p &= \text{SAGE}(F^s_{\mathcal{J}}, W^t_{\mathcal{J}, 1 \to q}, W^s_{\mathcal{J}, 1 \to q}) \\
I'_p &= \text{Average Pooling}(F^s_{1 \to \mathcal{J}}, I_p)
\end{aligned}
\end{equation}

\subsection{Speech Synthesizer}
We adopt the speech synthesizer with the same architecture as I$^3$-CSS \cite{jia2024intra}. Note that the feature aggregator of MFCIG-CSS adds the semantic interaction features $I'_s$ and prosodic interaction features $I'_p$ into $P_{t_{\mathcal{C}}}$ to constrain the synthesis of speech with conversational interaction prosody. The speech synthesis loss follows the setup of FastSpeech 2 \cite{ren2021fastspeech}.

\begin{table*}[t!]

\centering
\resizebox{1\linewidth}{!}{
\begin{tabular}{lccccc}
\toprule
\textbf{Systems}  & \textbf{N-DMOS} ($\uparrow$)  & \textbf{P-DMOS} ($\uparrow$)   & \textbf{MAE-P} ($\downarrow$)   & \textbf{MAE-E} ($\downarrow$)   & \textbf{MCD} ($\downarrow$)     \\ 
\midrule
Base-CTTS \cite{guo2021conversational}     & 3.673 $\pm$ 0.025    & 3.543 $\pm$ 0.027    & 0.530   & 0.467   & 11.42     \\

FCTalker \cite{hu2024fctalker}     & 3.716 $\pm$ 0.022    & 3.627 $\pm$ 0.021    & 0.479   & 0.325   & 11.41     \\

M$^2$-CTTS \cite{xue2023m}     & 3.756 $\pm$ 0.024    & 3.628 $\pm$ 0.028    & 0.543   & 0.380   & 11.96   \\

CONCSS \cite{deng2024concss}    & 3.819 $\pm$ 0.022    & 3.695 $\pm$ 0.024    & 0.482   & 0.328   & 11.92   \\

MSRGCN-CSS \cite{li2022inferring}    & 3.825 $\pm$ 0.020    & 3.734 $\pm$ 0.024    & 0.489   & 0.320   & 10.42   \\

ECSS \cite{liu2024emotion}     & 3.843 $\pm$ 0.022    & 3.770 $\pm$ 0.025    & 0.505   & 0.332   & 9.90   \\ 

I$^3$-CSS \cite{jia2024intra}     & 3.858 $\pm$ 0.022    & 3.795 $\pm$ 0.020    & 0.450   & \textbf{0.310}   & 11.47   \\ 

\midrule

\textbf{MFCIG-CSS (Proposed)}  & \textbf{3.980 $\pm$ 0.022} \textcolor[rgb]{0.0, 0.52, 0.52}{\tiny \( (+0.122) \)}   & \textbf{3.899 $\pm$ 0.024} \textcolor[rgb]{0.0, 0.52, 0.52}{\tiny \( (+0.104) \)}    & \textbf{0.439} \textcolor[rgb]{0.0, 0.52, 0.52}{\tiny \( (+0.011) \)}   & 0.314  & \textbf{9.53} \textcolor[rgb]{0.0, 0.52, 0.52}{\tiny \( (+0.37) \)}     \\ 

\bottomrule

\end{tabular}
}

\caption{Main results. Bold indicates the best result. Green indicates improvement over the best baseline. }

\label{tab1}
\end{table*}

\begin{table*}[t!]


\centering
\resizebox{1\linewidth}{!}{
\begin{tabular}{lccccc}
\toprule
\textbf{Systems}  & \textbf{N-DMOS} ($\uparrow$)  & \textbf{P-DMOS} ($\uparrow$)   & \textbf{MAE-P} ($\downarrow$)   & \textbf{MAE-E} ($\downarrow$)   & \textbf{MCD} ($\downarrow$)     \\ 
\midrule
Abl.Exp.1: w/o SIG     & 3.833 $\pm$ 0.025    & 3.793 $\pm$ 0.022    & 0.454   & 0.328   & 11.45     \\

Abl.Exp.2: w/o PIG     & 3.824 $\pm$ 0.025    & 3.765 $\pm$ 0.023    & 0.457   & 0.325   & 11.36   \\

Abl.Exp.3: w/o SIG and PIG   & 3.592 $\pm$ 0.023    & 3.512 $\pm$ 0.022    & 0.681   & 0.588   & 12.31   \\

\midrule

\textbf{MFCIG-CSS (Proposed)}  & \textbf{3.980 $\pm$ 0.022} \textcolor[rgb]{0.0, 0.52, 0.52}{\tiny \( (+0.147) \)}   & \textbf{3.899 $\pm$ 0.024} \textcolor[rgb]{0.0, 0.52, 0.52}{\tiny \( (+0.106) \)}    & \textbf{0.439} \textcolor[rgb]{0.0, 0.52, 0.52}{\tiny \( (+0.015) \)}   & \textbf{0.314} \textcolor[rgb]{0.0, 0.52, 0.52}{\tiny \( (+0.011) \)}  & \textbf{9.53} \textcolor[rgb]{0.0, 0.52, 0.52}{\tiny \( (+1.83) \)}     \\ 

\bottomrule

\end{tabular}
}

\caption{Ablation results. Bold indicates the best result. Green indicates improvement over the best ablation model.}

\label{tab2}

\end{table*}


\section{Experiments and Results}
\subsection{Dataset}
We validate MFCIG-CSS on the English dialogue dataset DailyTalk \cite{lee2023dailytalk}, which comprises 2,541 dialogue pairs with approximately 20 hours of speech data. Each dialogue consists of an average of 9.356 turns. The dialogues are recorded with alternating turns between a male and a female speaker. We split the data into training, validation, and test sets in an 8:1:1 ratio.




\subsection{Experiment Setup}
In MFCIG-CSS, the feature dimensions for text word-level, speech word-level, text utterance-level, and speech utterance-level in both $G_s$ and $G_p$ are set to 256. The SIG Encoder and PIG Encoder utilize SAGE \cite{hamilton2017inductive} for graph encoding, with both input and output channels set to 256. The speaker embedding dimension is also 256. The speech synthesizer configuration is based on FastSpeech2.0 \cite{ren2021fastspeech}. MFCIG-CSS is trained for 400k steps with a batch size of 16 on a single A800 GPU.

\subsection{Comparative and Ablation Models}
To demonstrate the effectiveness of the proposed MFCIG-CSS, we compare it with seven state-of-the-art CSS models. A detailed introduction of the compared models is provided in Appendix \ref{appendix:A1}.

For the ablation models, Abl.Exp.1 removes SIG to assess the impact of the semantic interaction graph; Abl.Exp.2 removes PIG to assess the impact of the prosody interaction graph; Abl.Exp.3 removes both to evaluate their combined effect on model performance.

\subsection{Evaluation Metric Details}
For subjective evaluation, we use the Dialogue-level Mean Opinion Score (DMOS) \cite{streijl2016mean, liu2024emphasis, LIU2025102948}. The evaluation is conducted by 20 graduate students specializing in speech, all of whom have passed CET-6, IELTS, or TOEFL exams and have extensive experience in DMOS assessments. Following the setup in \cite{jia2024intra}, we employ a 1-5 scale for Naturalness DMOS (N-DMOS) and Prosody DMOS (P-DMOS) to evaluate the quality and prosodic performance of the synthesized speech.

For objective evaluation, we compute the Mean Absolute Error of Pitch (MAE-P) and Mean Absolute Error of Energy (MAE-E) \cite{liu2024emotion} to assess the prosody of the synthesized speech. Additionally, we measure the Mel Cepstral Distortion (MCD) \cite{kubichek1993mel, chen2022v2c} between the synthesized and ground-truth speech to evaluate synthesis quality.

\subsection{Main Results}
We compare MFCIG-CSS with seven state-of-the-art CSS models and analyze the results. As shown in Table \ref{tab1}, MFCIG-CSS outperforms all baseline models in terms of average performance.
For subjective metrics, N-DMOS (3.980) and P-DMOS (3.899) achieve optimal performance, improving by 0.122 and 0.104 compared to the best baseline model, respectively. For objective metrics, MAE-P (0.439) and MCD (9.53) also achieve optimal performance, improving by 0.011 and 0.37 compared to the best baseline model. For the objective metric MAE-E (0.314), MFCIG-CSS achieves the second-best performance, just 0.004 lower than the best baseline model. The experimental results show that MFCIG-CSS, by explicitly modeling the interactions between word-level semantics, prosody, and the semantics, prosody of subsequent utterances in MDH, can better understand the conversational prosody expressed in MDH, thus enhancing the agent's ability to synthesize speech with appropriate conversational interaction prosody.

\subsection{Ablation Results}
To assess the contribution of each component in MFCIG-CSS, we conduct ablation experiments by removing different components, as shown in Table \ref{tab2}. Abl.Exp.1 removes SIG to verify the impact of modeling the interaction between word-level semantics, prosody, and the semantics of subsequent utterances in MDH on the performance of MFCIG-CSS. The experimental results show that removing SIG leads to a decrease in both subjective and objective metrics, indicating that explicitly modeling the semantic interaction in MDH with SIG helps improve the quality of the synthesized speech and enhances its conversational prosody. Abl.Exp.2 removes PIG to verify the impact of modeling the interaction between word-level semantics, prosody, and the prosody of subsequent utterances in MDH  on MFCIG-CSS performance. The experimental results show that removing PIG decreases all metrics, especially prosody-related metrics. This suggests that explicitly modeling the prosody interaction in MDH with PIG  helps the model learn the prosody interactions effectively, improving the conversational prosody of the synthesized speech. Abl. Exp. 3 removes both SIG and PIG, resulting in the worst performance across all metrics, further validating the significant contribution of SIG and PIG to the synthesis quality and prosody expressiveness of MFCIG-CSS.

\section{Conclusion}
To enhance the CSS system's understanding of MDH and enable the synthesis of speech with appropriate conversational prosody, we propose MFCIG-CSS, a novel framework that explicitly encodes the interactions between word-level semantics, prosody, and the semantics and prosody of subsequent utterances in MDH. This improves the model’s comprehension of both semantic and prosodic interactions within MDH. Experiments on DailyTalk demonstrate that MFCIG-CSS surpasses state-of-the-art CSS systems in prosody expression. In the future, we will explore the interaction modeling of finer-grained acoustic prosody, such as emotions, emphasis, pauses, etc., within MDH.

\section{Limitations}
One limitation of our work is that MFCIG-CSS is currently implemented only based on the FastSpeech 2 architecture to validate the effectiveness of the proposed semantic and prosody interaction graph modules. In the future, we plan to extend this approach to VITS-based architectures and discrete token-based speech encoders.
Another limitation is that we have not yet incorporated acoustic features such as emotion, emphasis, and pauses into the interaction graph modeling. Future work will focus on integrating these features to further enhance the prosodic expressiveness and naturalness of the synthesized speech.

\section{Acknowledgment}

The research by Rui Liu was funded by the Young Scientists Fund (No. 62206136), the General Program (No. 62476146) of the National Natural Science Foundation of China,  the Young Elite Scientists Sponsorship Program by CAST (2024QNRC001), the Outstanding Youth Project of Inner Mongolia Natural Science Foundation (2025JQ011), Key R\&D and Achievement Transformation Program of Inner Mongolia Autonomous Region (2025YFHH0014) and the Central Government Fund for Promoting Local Scientific and Technological Development (2025ZY0143). The work of Zhenqi Jia was funded by the Research and Innovation Project for Graduate Students of Inner Mongolia University. The work by Berrak Sisman was supported by NSF CAREER award IIS-2338979. The work by Haizhou Li was supported by the Shenzhen Science and Technology Program (Shenzhen Key Laboratory, Grant No. ZDSYS20230626091302006), the Shenzhen Science and Technology Research Fund (Fundamental Research Key Project, Grant No. JCYJ20220818103001002), and the Program for Guangdong Introducing Innovative and Enterpreneurial Teams, Grant No. 2023ZT10X044.

\bibliography{emnlp}

\appendix

\section{Example Appendix}
\label{sec:appendix}

\subsection{Comparative Models}
\label{appendix:A1}
\begin{itemize}
    \item \textbf{Base-CTTS \cite{guo2021conversational}} introduces a text coarse-grained context encoder to improve the quality of synthesized speech. 
    
    \item \textbf{FCTalker \cite{hu2024fctalker}} designs a coarse-grained and fine-grained text context encoder to enhance the prosody of synthesized speech. 
    
    \item \textbf{M$^2$-CTTS \cite{xue2023m}} proposes a multi-scale, multi-modal context encoder to enhance the prosody of synthesized speech. 
    
    \item \textbf{CONCSS \cite{deng2024concss}} incorporates a negative sample enhancement sampling strategy in MDH modeling to improve the prosody sensitivity of synthesized speech. 
    
    \item \textbf{MSRGCN-CSS \cite{li2022inferring}} introduces a context modeling scheme based on a multi-scale relational graph convolutional network to enhance the speaking style of synthesized speech. 
    
    \item \textbf{ECSS \cite{liu2024emotion}} incorporates a context modeling scheme based on multi-source knowledge heterogeneous graphs to enhance the emotional expressiveness of the synthesized speech.  
    
    \item \textbf{I$^3$-CSS \cite{jia2024intra}} includes an intra-modal and inter-modal context interaction modeling scheme at the utterance level to improve the prosody performance of synthesized speech. 
\end{itemize}

\end{document}